\documentclass{article}
\usepackage{spconf,amsmath,graphicx}

\usepackage{graphicx}
\usepackage{booktabs}
\usepackage{amsmath,amsfonts,amssymb}
\usepackage{multirow}
\usepackage{algorithm}
\usepackage{algorithmic}
\usepackage{subfigure}
\usepackage{amsthm}
\usepackage{url}            
\usepackage{amsfonts}       
\usepackage{amsmath}
\newcommand{\xvec}{{\bf x}}

\title{Maximum Batch Frobenius Norm for Multi-Domain Text Classification}
\name{Yuan Wu$^{\star}$ \qquad Diana Inkpen$^{\dagger}$ \qquad Ahmed El-Roby$^{\star}$}
\address{$^{\star}$ Carleton University, Ottawa, Canada \\
      $^{\dagger}$ University of Ottawa, Ottawa, Canada}
\begin{document}
%
\maketitle
\begin{abstract}
Multi-domain text classification (MDTC) has obtained remarkable achievements due to the advent of deep learning. Recently, many endeavors are devoted to applying adversarial learning to extract domain-invariant features to yield state-of-the-art results. However, these methods still face one challenge: transforming original features to be domain-invariant distorts the distributions of the original features, degrading the discriminability of the learned features. To address this issue, we first investigate the structure of the batch classification output matrix and theoretically justify that the discriminability of the learned features has a positive correlation with the Frobenius norm of the batch output matrix. Based on this finding, we propose a maximum batch Frobenius norm (MBF) method to boost the feature discriminability for MDTC. Experiments on two MDTC benchmarks show that our MBF approach can effectively advance the performance of the state-of-the-art.
\end{abstract}
\section{Introduction}
\label{sec:intro}

Text classification is a fundamental task in natural language processing (NLP) \cite{minaee2021deep}. In recent years, text classification models have demonstrated significant improvements by employing deep neural networks (DNNs) \cite{goodfellow2016deep}. These achievements mainly rely on large amounts of labeled data. Unfortunately, in many real-world applications, the availability of labeled data may vary across different domains \cite{pan2009survey}. For domains like book and movie reviews, abundant labeled data has been made available, while for domains like medical equipment reviews, labeled data is scarce. Therefore, it is of great significance to investigate how to improve the classification accuracy on the target domain by leveraging available resources from related domains.

Multi-domain text classification (MDTC) is proposed to address the above issue. Recent advances in MDTC are reliant on adversarial learning to enhance the transferability of feature representations \cite{ganin2016domain}, and shared-private paradigm to benefit the system performance from domain-specific knowledge \cite{bousmalis2016domain}. Adversarial learning is first proposed by generative adversarial networks (GANs) for image generation \cite{goodfellow2014generative}. It uses noise data to generate images and enforce the distribution of the generated images to be similar to that of the real images. The success of GANs inspires the adversarial alignment \cite{liu2017adversarial,chen2018multinomial,wu2021towards}. Adversarial alignment can harness the power of DNNs to learn domain-invariant features that are both discriminative and transferable by deploying a minimax game between a feature extractor and a domain discriminator: the discriminator aims to distinguish features across different domains, while the feature extractor contrives to deceive the discriminator. When these two components reach equilibrium, the learned features are regarded as domain-invariant. As text classification is a well-known domain-specific task \cite{aggarwal2012survey}, the same word in different domains may express different sentiments. For instance, the word \textit{fast} represents negative sentiment in the electronic review "The battery of the camera runs fast", while it indicates positive sentiment in the car review "The car runs fast". Domain-specific knowledge cannot be ignored in MDTC. Shared-private paradigm is proposed to incorporate domain-specific features to enhance the discriminability of the domain-invariant features \cite{chen2018multinomial}. The MDTC models that adopt adversarial learning and shared-private paradigm can yield state-of-the-art performance \cite{wu2021conditional,wu2021mixup}. However, these models still face one critical issue: as adversarial alignment needs to distort the original feature distributions to generate domain-invariant features, the transferability of the domain-invariant feature is obtained at the expense of sacrificing the discriminability \cite{liu2019transferable}. 

In this paper, we address the aforementioned issue by proposing a maximum batch Frobenius norm (MBF) approach to boost the feature discriminability. We first investigate the structure of the classification output matrix of a randomly sampled data batch, and then theoretically analyze how the discriminability of the learned features can be measured by the batch classification output matrix. We find that the feature discriminability has a positive correlation with the Frobenius norm of the batch output matrix. Motivated by this finding, our MBF method boosts the feature discriminability by maximizing the Frobenius norm of the batch output matrix. Experiments are conducted on two MDTC benchmarks: the Amazon review dataset and the FDU-MTL dataset. The experimental results show that our MBF method can outperform the state-of-the-art baselines on both datasets.

\section{Proposed Method}
\label{sec:method}

In this paper, we study the MDTC problem, where the text data comes from multiple domains, each with varying amounts of both labeled and unlabeled data. Assume we have $M$ different domains $\{D_i\}_{i=1}^M$, the $i$-th domain $D_i$ contains a limited amount of labeled data $\mathbb{L}_i=\{(\xvec_j,y_j)\}_{j=1}^{l_i}$ and a large amount of unlabeled data $\mathbb{U}_i=\{\xvec_j\}_{j=1}^{u_i}$, where $l_i$ and $u_i$ are the numbers of labeled and unlabeled data in $D_i$. Our goal is to learn a model that maps an instance $\xvec$ to its corresponding label $y$ and guarantee the model generalizes well over all domains. 

\subsection{Adversarial Multi-Domain Text Classification}

Adversarial MDTC methods, starting from adversarial multi-task learning for text classification (ASP-MTL) \cite{liu2017adversarial}, have become increasingly influential in MDTC. The main idea is to adversarially align different feature distributions to extract both explanatory and transferable features. Basically, the standard MDTC model has four components: a shared feature extractor $\mathcal{F}_s$, $M$ domain-specific feature extractors $\{\mathcal{F}_d^i\}_{i=1}^M$, a domain discriminator $\mathcal{D}$, and a classifier $\mathcal{C}$. The shared feature extractor $\mathcal{F}_s$ learns domain-invariant features, while the $i$-th domain-specific feature extractor $\mathcal{F}_d^i$ captures domain-specific knowledge of the $i$-th domain $D_i$. The two types of feature extractors are supposed to complement each other and maximally capture useful information across domains. The discriminator $\mathcal{D}_i$ is a $M$-class classifier which takes a domain-invariant feature vector as input and outputs the probability vector for input data; i.e., $\mathcal{D}_i(\mathcal{F}_s(\xvec))$ denotes the probability of instance $\xvec$ coming from the $i$-th domain. In this paper, we use the negative log-likelihood (NLL) loss to encode the multinomial adversarial loss:

\begin{align}
    \mathcal{L}_{Adv}=-\sum_{i=1}^M\mathbb{E}_{\xvec\sim\mathbb{L}_i\cup\mathbb{U}_i}\log(\mathcal{D}_i(\mathcal{F}_s(\xvec)))
\end{align}

By training $\mathcal{F}_s$ adversarially to deceive $\mathcal{D}$, the learned features are made transferable across domains. In addition, the classifier $\mathcal{C}$ are trained together with $\mathcal{F}_s$ and $\{\mathcal{F}_d^i\}_{i=1}^M$ to minimize the classification error on the labeled data. This makes the feature representations discriminative across categories. The classifier $\mathcal{C}$ takes the concatenation of a domain-invariant feature and a domain-specific feature as input, and predicts the label probability; i.e., $\mathcal{C}_y([\mathcal{F}_s(\xvec),\mathcal{F}_d^i(\xvec)])$ denotes the probability of instance $\xvec$ belonging to label $y$, where $[\cdot,\cdot]$ represents the concatenation of two vectors. We also use NLL loss to formulate the classification loss:

\begin{align}
    \mathcal{L}_c=-\sum_{i=1}^M\mathbb{E}_{(\xvec,y)\sim\mathbb{L}_i}\log(\mathcal{C}_y([\mathcal{F}_s(\xvec),\mathcal{F}_d^i(\xvec)]))
\end{align}

\subsection{Frobenius Norm of Batch Output Matrix}

DNNs are often trained by using mini-batch stochastic gradient descent (SGD). Here, we set the batch size to $B$ and assume we handle the $K$-class classification. Then the batch output matrix can be represented as $A\in\mathbb{R}^{B\times K}$. The matrix $A$ should satisfy two conditions: (1) the sum of each entry of A is $1$; (2) each entry is non-negative, i.e.:

\begin{equation}
    \begin{aligned}
    \sum_{j=1}^K&A_{i,j}=1\quad\forall i\in 1...B \\
    &A_{i,j}\geq0\quad\forall i\in 1...B,j\in1...K
    \end{aligned}
\end{equation}

In standard supervised machine learning, a well-performed model could learn features with high discriminability by processing abundant labeled data. While in label-scarce scenarios, limited amounts of labeled data can not guarantee the sufficient discriminability of the learned features and the data without supervision often deteriorate the feature discriminability \cite{zhu2009introduction}. Therefore, to learn qualified features in MDTC, we need to optimize the predictions on unlabeled data.

Semi-supervised learning (SSL) is a typical label-scarce machine learning task. In SSL, entropy minimization is often used to minimize the uncertainty of prediction on unlabeled data \cite{grandvalet2005semi}. In general, lower uncertainty suggests higher discriminability. The entropy of matrix $A$ is formulated as:

\begin{align} \label{eq4}
    H(A)=-\frac{1}{B}\sum_{i=1}^B\sum_{j=1}^KA_{i,j}\log(A_{i,j})
\end{align}

Inspired by entropy minimization \cite{grandvalet2005semi}, stronger discriminability is often achieved by minimizing $H(A)$. When $H(A)$ is optimized to its minimum, for each row of $A$, only one entry is $1$ and the other $K-1$ entries are $0$. The minimal $H(A)$ leads to the highest discriminability, where each item $A_i$ is fully determined.

We denote the Frobenius norm of the matrix $A$ as $\|A\|_{F}$. As $\|A\|_{F}$ has strict opposite monotonicity with $H(A)$ (See section \ref{thm} for the theoretical justification), maximizing $\|A\|_{F}$ is equivalent to minimizing $H(A)$. Therefore, the feature discriminability can be enhanced by maximizing $\|A\|_{F}$. $\|A\|_{F}$ is formulated as:

\begin{align} \label{eq5}
    \|A\|_{F}=\sqrt{\sum_{i=1}^B\sum_{j=1}^K|A_{i,j}|^2}
\end{align}

To improve the discriminability of the learned features, we maximize the Frobenius norm of the batch classification output matrix of the unlabeled data. The corresponding loss function is formulated as:

\begin{align}
    \mathcal{L}_{BF}=\sum_{i=1}^M\mathbb{E}_{\xvec\sim\mathbb{U}_i}\frac{1}{B}\|\mathcal{C}^B([\mathcal{F}_s(\xvec),\mathcal{F}_d^i(\xvec)])\|_{F}
\end{align}

\noindent where $\mathcal{C}^B(\cdot)$ represents the batch output matrix with $B$ randomly sampled unlabeled data. Maximizing $\mathcal{L}_{BF}$ could improve the feature discriminability without degrading the transferability, the gradient of Frobenius norm could be calculated according to \cite{papadopoulo2000estimating}. Therefore, $\mathcal{L}_{BF}$ could be applied to optimize the gradient-based DNNs.

\subsection{Maximum Batch Frobenius Norm}

Based on the above finding, we propose the maximum batch Frobenius norm (MBF) method, whose objective function can be written as:

\begin{align}
    \min_{\mathcal{F}_s,\{\mathcal{F}_d^i\}_{i=1}^M,\mathcal{C}}\max_{\mathcal{D}}\mathcal{L}_c+\alpha\mathcal{L}_{Adv}-\beta\mathcal{L}_{BF}
\end{align}

\noindent where $\alpha$ and $\beta$ are hyperparameters that balance different loss functions. The MBF model can be trained with mini-batch SGD and we adopt the alternating fashion \cite{goodfellow2014generative} to achieve the minimax optimization. The training algorithm is presented in Algorithm \ref{trainingalg}.

\begin{algorithm}
\caption{SGD training algorithm}\label{trainingalg}
\begin{algorithmic}[1]
\STATE{\bf Input:} labeled data $\mathbb{L}_i$ and unlabeled data $\mathbb{U}_i$ in $M$ domains; 
	two hyperparameters $\alpha$ and $\beta$.
\FOR{number of training iterations}
	\STATE Sample labeled mini-batches from the multiple domains $B^\ell=\{B^\ell_1,\cdots, B^\ell_M\}$.
	\STATE Sample unlabeled mini-batches from the multiple domains $B^u=\{B^u_1,\cdots, B^u_M\}$.
	\STATE Calculate $loss = \mathcal{L}_{\mathcal{C}}+\alpha \mathcal{L}_{Adv}-\beta\mathcal{L}_{BF}$ on $B^\ell$ and $B^u$;\\
	Update $\mathcal{F}_s$, $\{\mathcal{F}_d^i\}_{i=1}^M$, $\mathcal{C}$ by descending along the gradients $\Delta loss$.\\[.2ex] 
	\STATE Calculate $l_D=\mathcal{L}_{Adv}$ on $B^\ell$ and $B^u$;\\
	Update $\mathcal{D}$ by ascending along the gradients $\Delta l_D$.\\[.2ex]
\ENDFOR
\end{algorithmic}
\end{algorithm}

\subsection{Theoretical Analysis} \label{thm}

As shown in Eq.\ref{eq5}, the Frobenius norm of matrix $A$ is the square root sum of all entries in $A$. The calculating process can be divided into two steps: (1) calculating the quadratic sum of each row in $A$; (2) calculating the square root of the sum of all rows. As the monotonicity of the square root of the sum of all rows depends on the monotonicity of each row and there is no extra constraint on different rows, we consider the monotonicity of the quadratic sum of each row to analyze the monotonicity of the Frobenius norm. Analogously, we could also draw the monotonicity of the entropy by analyzing the monotonicity of each row for the entropy.

We use $f(A_i)$ to denote the square sum of the $i$-th row in $A$, so we have $f(A_i)=\sum_{j=1}^KA_{i,j}^2$. To analyze the monotonicity of a function on multiple variables, we could simply analyze the monotonicity of the function on each variable. We make the assumption that the variable $A_{i,K}$ is the only variable dependent on $A_{i,j}$ as the variables are supposed to be independent and $\sum_{j=1}^KA_{i,j}=1$. Then the partial derivative of $f(A_i)$ can be calculated as:

\begin{equation}
    \begin{aligned}
    \frac{\partial f(A_i)}{\partial A_{i,j}}=2A_{i,j}-2A_{i,K}=4A_{i,j}-2(1-\sum_{k=1,k\neq j}^{K-1}A_{i,k})
    \end{aligned}
\end{equation}

\noindent We can observe that when $A_{i,j}\leq\frac{1}{2}-\frac{1}{2}\sum_{k=1,k\neq j}^{K-1}A_{i,k}$, $f(A_i)$ decreases monotonously. while $A_{i,j}\geq\frac{1}{2}-\frac{1}{2}\sum_{k=1,k\neq j}^{K-1}A_{i,k}$, $f(A_{i})$ increases monotonously.

Based on Eq.\ref{eq4}, we define the entropy of the $i$-th row in A as:

\begin{align}
    h(A_i)=-\sum_{j=1}^KA_{i,j}\log(A_{i,j})
\end{align}

\noindent Furthermore, the partial derivative of $h(A_i)$ can be calculated as:

\begin{equation}\label{eq10}
    \begin{aligned}
    \frac{\partial h(A_i)}{\partial A_{i,j}}&=-\log(A_{i,j})+\log(A_{i,K}) \\
    &=\log(\frac{1-A_{i,j}-\sum_{k=1,k\neq j}^{K-1}A_{i,k}}{A_{i,j}})
    \end{aligned}
\end{equation}

\noindent From Eq.\ref{eq10}, we can observe that when $A_{i,j}\leq\frac{1}{2}-\frac{1}{2}\sum_{k=1,k\neq j}^{K-1}A_{i,k}$, $h(A_i)$ increases monotonously. While $A_{i,j}\geq\frac{1}{2}-\frac{1}{2}\sum_{k=1,k\neq j}^{K-1}A_{i,k}$, $h(A_i)$ decreases monotonously. In summary, we draw the conclusion that the Frobenius norm and entropy of a matrix have strict opposite monotonicity.

\section{Experiments}
\label{sec:experiment}

\subsection{Dataset}

We evaluate our MBF method on two MDTC benchmarks: the Amazon review dataset \cite{blitzer2007biographies} and the FDU-MTL dataset \cite{liu2017adversarial}. The Amazon review dataset contains four domains: books, dvds, electronics, and kitchen. Each domain has 1,000 positive samples and 1,000 negative samples. All data have been pre-processed into a bag of features (unigrams and bigrams), losing word order information. We thus take the 5,000 most frequent features such that each review is encoded as a 5,000-dimensional vector, where feature values are raw counts of the features. The FDU-MTL dataset is a more challenging dataset, containing 16 domains: 14 product review domains (books, electronics, DVDs, kitchen, apparel, camera, health, music, toys, video, baby, magazine, software, and sport) and 2 movie review domains (IMDB and MR). The reviews in this dataset are raw text data only being tokenized by the Stanford tokenizer \cite{manning2014stanford}. Each domain has a development set of 200 samples and a test set of 400 samples. The amounts of training and unlabeled data vary among domains but are roughly 1,400 and 2,000. 

\subsection{Baselines}

We compare our MBF approach with several baselines. The multi-task deep neural network model (MT-DNN) uses a shared layer for different domains and bag-of-words input \cite{liu2015representation}. The collaborative multi-domain sentiment classification (CMSC) model uses the shared-private paradigm and can be trained by three loss functions: the least square loss (CMSC-L2), the hinge loss (CMSC-SVM), and the log loss (CMSC-Log) \cite{wu2015collaborative}. The adversarial multi-task learning for text classification (ASP-MTL) combines adversarial learning, shared-private paradigm, and orthogonality regularizer to train models \cite{liu2017adversarial}. The multinomial adversarial network (MAN) tries to encode the domain discrepancy by two f-divergences: the negative log-likelihood loss (MAN-NLL) and the least square loss (MAN-L2) \cite{chen2018multinomial}. The dual adversarial Co-learning (DACL) method uses discriminator-based adversarial learning and classifier-based adversarial learning to learn domain-invariant features \cite{wu2020dual}. The conditional adversarial network (CAN) aligns joint distributions of shared features and label predictions to optimize the adversarial alignment \cite{wu2021conditional}. 

\subsection{Implementation Details}

We follow the standard MDTC setting and adopt the same network architectures with the most recent baselines for fair comparisons \cite{chen2018multinomial,wu2020dual,wu2021conditional}. For the Amazon review dataset, we employ multi-layer perceptron (MLP) with an input size of 5,000 as feature extractors. Each feature extractor has two hidden layers with sizes 1,000 and 500. The output sizes of $\mathcal{F}_s$ and $\{\mathcal{F}_d^i\}_{i=1}^M$ are 128 and 64, respectively. $\mathcal{C}$ and $\mathcal{D}$ are also MLPs with one hidden layer of the same size as its input ($128+64$ for $\mathcal{C}$ and $128$ for $\mathcal{D}$). We conduct 5-fold cross-validation on this dataset and report the 5-fold average results. For the FDU-MTL dataset, a one-layer convolutional neural network (CNN) is used as the feature extractor. The CNN uses different kernel sizes $(3,4,5)$, and the number of kernels is 200. The input of the CNN is a 100-dimensional embedding obtained by processing each word of the input sequence through word2vec \cite{mikolov2013efficient}. For all experiments, we set $\alpha=0.5$ and $\beta=1$ and take the convenience to cite the experimental results directly from \cite{wu2020dual,wu2021conditional}.

\subsection{Results}

\begin{table}
  \caption{MDTC results on the Amazon review dataset}
  \label{table_ref1}
  \resizebox{1.0\columnwidth}{!}{
  \begin{tabular}{l| c c c c c c c c}
    \toprule
    Domain & CMSC-LS & CMSC-SVM & CMSC-Log & MAN-L2 & MAN-NLL & DACL & CAN & MBF(Proposed)\\
    \midrule
    Books &  82.10 & 82.26 & 81.81 & 82.46 & 82.98 & 83.45 & 83.76 & \textbf{84.58$\pm$0.21} \\
    DVDs &  82.40 & 83.48 & 83.73 & 83.98 & 84.03 & 85.50 & 84.68 & \textbf{85.78$\pm$0.07} \\
    Electr.  & 86.12 & 86.76 & 86.67 & 87.22 & 87.06 & 87.40 & 88.34 & \textbf{89.04$\pm$0.11} \\
    Kit.  &  87.56 & 88.20 & 88.23 & 88.53 & 88.57 & 90.00 & 90.03 & \textbf{91.45$\pm$0.19}\\
    \midrule
    AVG  &  84.55 & 85.18 & 85.11 & 85.55 & 85.66 & 86.59 & 86.70 & \textbf{87.71$\pm$0.10}\\
    \bottomrule
  \end{tabular}}
\end{table}

\begin{table}
  \caption{MDTC results on the FDU-MTL dataset}
  \label{table_ref2}
  \resizebox{1.0\columnwidth}{!}{
  \begin{tabular}{l| c c c c c c c }
    \toprule
    Domain & MT-DNN & ASP-MTL & MAN-L2 & MAN-NLL & DACL & CAN & MBF(Proposed)\\
    \midrule
    books & 82.2 & 84.0 & 87.6 & 86.8 & 87.5 & 87.8 & \textbf{89.1$\pm$0.1} \\
    electronics & 81.7 & 86.8 & 87.4 & 88.8 & 90.3 & \textbf{91.6} & 91.0$\pm$0.7\\
    dvd & 84.2 & 85.5 & 88.1 & 88.6 & 89.8 & 89.5 & \textbf{90.4$\pm$0.2} \\
    kitchen & 80.7 & 86.2 & 89.8 & 89.9 & 91.5 & 90.8 & \textbf{93.3$\pm$0.7}\\
    apparel & 85.0 & 87.0 & 87.6 & 87.6 & \textbf{89.5} & 87.0 & 88.5$\pm$0.6 \\
    camera & 86.2 & 89.2 & 91.4 & 90.7 & 91.5 & \textbf{93.5} & 92.8$\pm$0.8\\
    health & 85.7 & 88.2 & 89.8 & 89.4 & 90.5 & 90.4 & \textbf{92.0$\pm$0.1} \\
    music & 84.7 & 82.5 & 85.9 & 85.5 & 86.3 & \textbf{86.9} & 85.9$\pm$0.4 \\
    toys & 87.7 & 88.0 & 90.0 & 90.4 & 91.3 & 90.0 & \textbf{92.2$\pm$0.5} \\
    video & 85.0 & 84.5 & 89.5 & 89.6 & 88.5 & 88.8 & \textbf{90.4$\pm$0.7} \\
    baby & 88.0 & 88.2 & 90.0 & 90.2 & \textbf{92.0} & \textbf{92.0} & 90.8$\pm$0.2\\
    magazine & 89.5 & 92.2 & 92.5 & 92.9 & 93.8 & \textbf{94.5} & 93.5$\pm$0.2 \\ 
    software & 85.7 & 87.2 & 90.4 & 90.9 & 90.5 & 90.9 & \textbf{91.4$\pm$0.4} \\
    sports & 83.2 & 85.7 & 89.0 & 89.0 & 89.3 & \textbf{91.2} & 90.3$\pm$0.1 \\
    IMDB & 83.2 & 85.5 & 86.6 & 87.0 & 87.3 & 88.5 & \textbf{89.9$\pm$0.1}\\
    MR & 75.5 & 76.7 & 76.1 & 76.7 & 76.0 & 77.1 & \textbf{79.2$\pm$0.5}\\
    \midrule
    AVG & 84.3 & 86.1 & 88.2 & 88.4  & 89.1  & 89.4 & \textbf{90.1$\pm$0.2} \\
    \bottomrule
  \end{tabular}}
\end{table}

All results are reported on five random trials. The experimental results on the Amazon review dataset are shown in Table \ref{table_ref1}. We can observe that the MBF method can achieve the best average classification accuracy of $87.71\%$, outperforming the second-best model CAN by a margin of $1.01\%$. Moreover, our approach can also beat other baselines on each individual domain. For domain DVDs and kitchen, our MBF method can improve the state-of-the-art by more than $1\%$. The performance difference between MBF and CAN is more notable than that between CAN and DACL. Table \ref{table_ref2} presents the results on the FDU-MTL dataset. It can be noted that the MBF provides the best performance comparing with other baselines, outperforming the CAN model by a margin of $0.7\%$. In particular, our method can obtain the best performance on 9 out of 16 domains. For domain books, kitchen, health, IMDB, and MR domains, we can advance the state-of-the-art performance by more than $1\%$.

\section{Conclusion}
\label{sec:conclusion}

In this paper, we propose a maximum batch Frobenius norm (MBF) method for MDTC. We find that the discriminability of the learned features can be measured by the Frobenius norm of the batch classification output matrix, and maximizing the Frobenius norm of the batch output matrix can reduce the uncertainty of prediction on unlabeled data, boosting the feature discriminability. Therefore, we use MBF to enhance the discriminability of the learned features without degrading the transferability such that the system performance can be improved. The extensive experiments on two MDTC benchmarks show that our MBF approach effectively advances state-of-the-art performance.

\bibliographystyle{IEEEbib}
\bibliography{strings}

\end{document}